\documentclass[10pt,twocolumn,letterpaper]{article}

\usepackage{iccv}              % To produce the CAMERA-READY version
\usepackage{caption}
\newcommand\blfootnote[1]{%
  \begingroup
  \renewcommand\thefootnote{}\footnote{#1}%
  \addtocounter{footnote}{-1}%
  \endgroup
}
\usepackage[accsupp]{axessibility}  % Improves PDF readability for those with disabilities.
\definecolor{iccvblue}{rgb}{0.21,0.49,0.74}
\usepackage[pagebackref,breaklinks,colorlinks,allcolors=iccvblue]{hyperref}

\def\paperID{*****} % *** Enter the Paper ID here
\def\confName{ICCV}
\def\confYear{2025}

\title{Fitting Image Diffusion Models on Video Datasets}

\author{Juhun Lee\\
Dept. of Artificial Intelligence\\
Sungkyungwan University\\
{\tt\small josejhlee@g.skku.edu}
\and
Simon S. Woo$^{\star}$\\
Dept. of Artificial Intelligence\\
Sungkyungwan University\\
{\tt\small swoo@g.skku.edu}
}

\begin{document}
\maketitle
\begin{abstract}

Image diffusion models are trained on independently sampled static images. While this is the bedrock task protocol in generative modeling, capturing the temporal world through the lens of static snapshots is information-deficient by design. This limitation leads to slower convergence, limited distributional coverage, and reduced generalization. In this work, we propose a simple and effective training strategy that leverages the temporal inductive bias present in continuous video frames to improve diffusion training. Notably, the proposed method requires no architectural modification and can be seamlessly integrated into standard diffusion training pipelines. We evaluate our method on the HandCo dataset, where hand-object interactions exhibit dense temporal coherence and subtle variations in finger articulation often result in semantically distinct motions. Empirically, our method accelerates convergence by over 2$\text{x}$ faster and achieves lower FID on both training and validation distributions. It also improves generative diversity by encouraging the model to capture meaningful temporal variations. We further provide an optimization analysis showing that our regularization reduces the gradient variance, which contributes to faster convergence.

\blfootnote{
$^\star$ Corresponding author
}

\end{abstract}

\section{Introduction}
\label{sec:intro}

Understanding and advancing the capabilities of diffusion models hinges, in large part, on how effectively they are fitted to data. To date, the community’s dominant focus for such study has been static image datasets, and success on this modality has become a cornerstone benchmark for theoretical and empirical progress in visual generative modeling.

% \begin{figure}[t]
%     \centering
%     \includegraphics[width=1.0\linewidth]{figure/qulative_results_fig.png}
%     \caption{Qualitative comparison of generated images. Left: samples from the baseline model. Right: samples generated by our ($\pi^{flow}$ variant) method. Our model produces more diverse outputs.
%     }
%     \label{fig:sample}
% \end{figure}

\begin{figure}[t]
    \centering
    \includegraphics[width=1.0\linewidth]{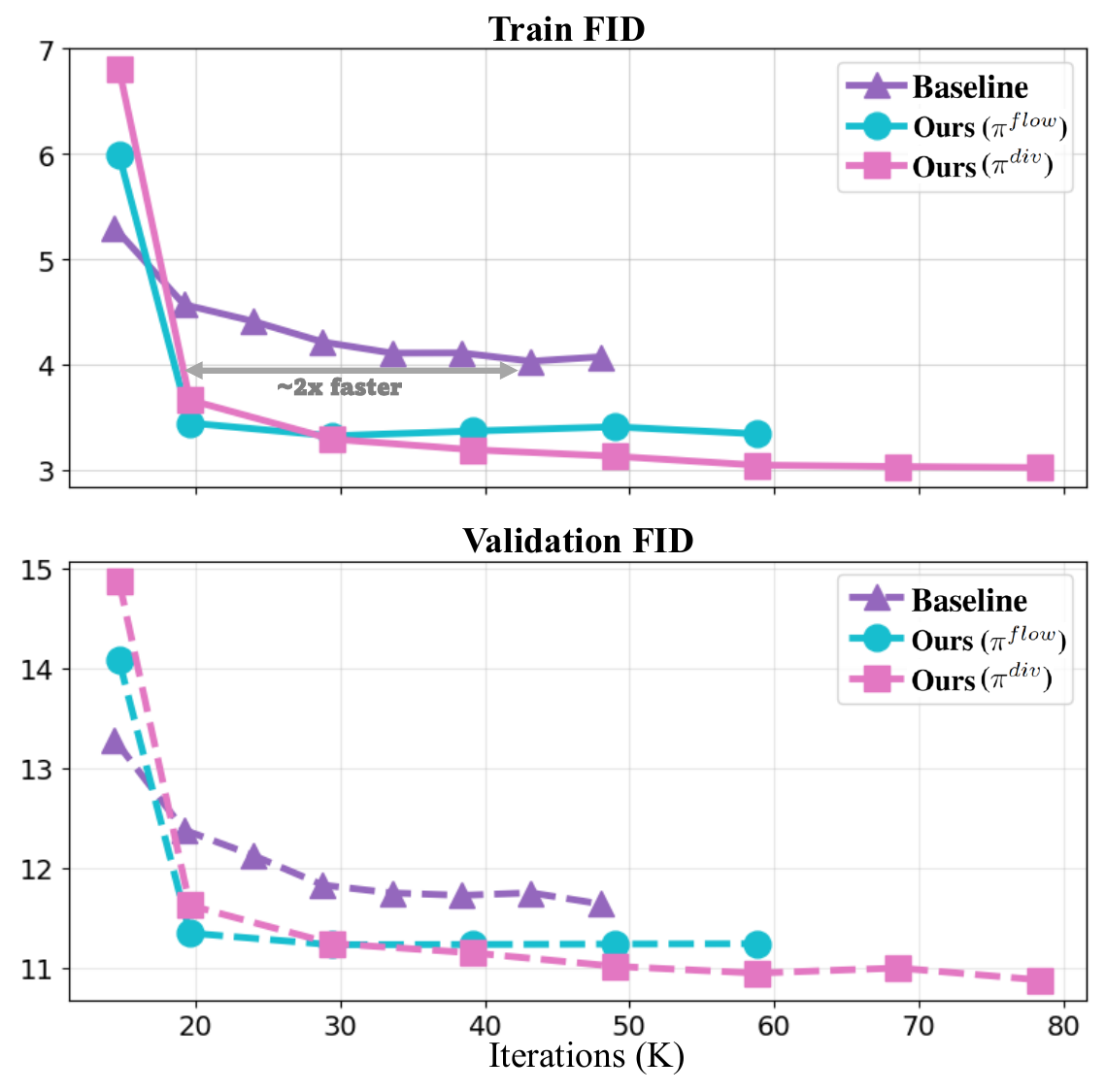}
    \caption{FID scores over training iterations for the baseline and our method variants. Our approach converges faster and consistently achieves lower train and validation FID, demonstrating improved training efficiency and generalization performance.
    }
    \label{fig:convergence}
\end{figure}

In this work we deliberately depart from that convention and examine an alternative training configuration: training image diffusion models on temporal (video) datasets. There are two principal motivations for embracing this configuration. 

Firstly, when every frame in a video is itself a valid and semantically meaningful image, the temporal stream forms a semantic traversal through the underlying data manifold. A hand-gesture dataset, for example, yields a sequence in which even the slightest articulation encodes a meaningful shift in semantics. By treating each frame as a legitimate training instance while simultaneously contextualizing it with its temporal neighbors, we can expose the model to a structured prior that is entirely absent when frames are shuffled into an i.i.d. pool. In effect, static snapshots of the world distribution are now embedded in a coherent local neighbourhood that supplies the model with auxiliary cues about continuity, shape, and appearance variation. 

Another motivation is rooted in training stability and convergence behavior. Diffusion models are notorious for slow convergence, and under data-scarce regimes, they often over-fit to a subset of “easy’’ modes. This pathology arises from an intrinsic coupling between the denoising objective and the accumulation of sampling error: a perfect denoiser would, in principle, memorise the training data, whereas, in practice, the small errors incurred at each reverse diffusion process step both ``bless'' the model with the ability to generate novel samples and ``doom'' it to preferentially denoise towards most readily reconstructible samples. Prior successes on data augmentation in GAN training (e.g.\ ADA, Diff-Aug, negative augmentation) suggest data augmentation can regularise diffusion training, temper over-fitting, and foster richer representation learning.

For these reasons, our work quantifies the influence of the data geometry presented during training and treats it as a controllable variable for both optimisation stability and convergence optimality. Concretely, we introduce a proximity-weighted Laplacian smoothing term that penalises discrepancies between predictions of temporally adjacent noisy inputs. To even our surprise, this single regularization yields markedly faster convergence (2$\text{x}$) and superior generative quality (a whole -1.0 train FID) on the training set, while simultaneously improving generalisation (-0.87 val. FID) beyond mere memorisation of seen data. Furthermore, our optimization analysis reveals that the regulariser reduces gradient variance by aligning local Jacobians, thereby explaining its stabilising effect. We hope that this study inspires further exploration of the data set structure as a lever to shape the training dynamics of diffusion models.

\section{Related Works}

Denoising diffusion probabilistic models (DDPMs) \cite{croitoru2023diffusion,ho2020denoising,dhariwal2021diffusion} have emerged as a powerful class of generative models, achieving state-of-the-art performance in image synthesis, editing, and conditional generation tasks. These models learn to reverse a gradual noising process through iterative denoising steps, guided by a neural network trained to predict added noise~\cite{ho2020denoising, dhariwal2021diffusion}. Variants have explored architectural improvements~\cite{nichol2021improved} and applications to various modalities such as text \cite{zhang2023adding}, video \cite{Hoogeboom23,Singer22,Wang24}. Complementary to these advances, recent work has sought to accelerate the notoriously long training cycles of diffusion models: the Min‑SNR weighting strategy~\cite{hang2023efficient} adaptively prioritizes low‑signal‑to‑noise training examples, while FasterDiT~\cite{yao2024fasterdit} introduces architectural and algorithmic refinements that markedly shorten convergence time.

\begin{figure}[t]
    \centering
    \includegraphics[width=1.0\linewidth]{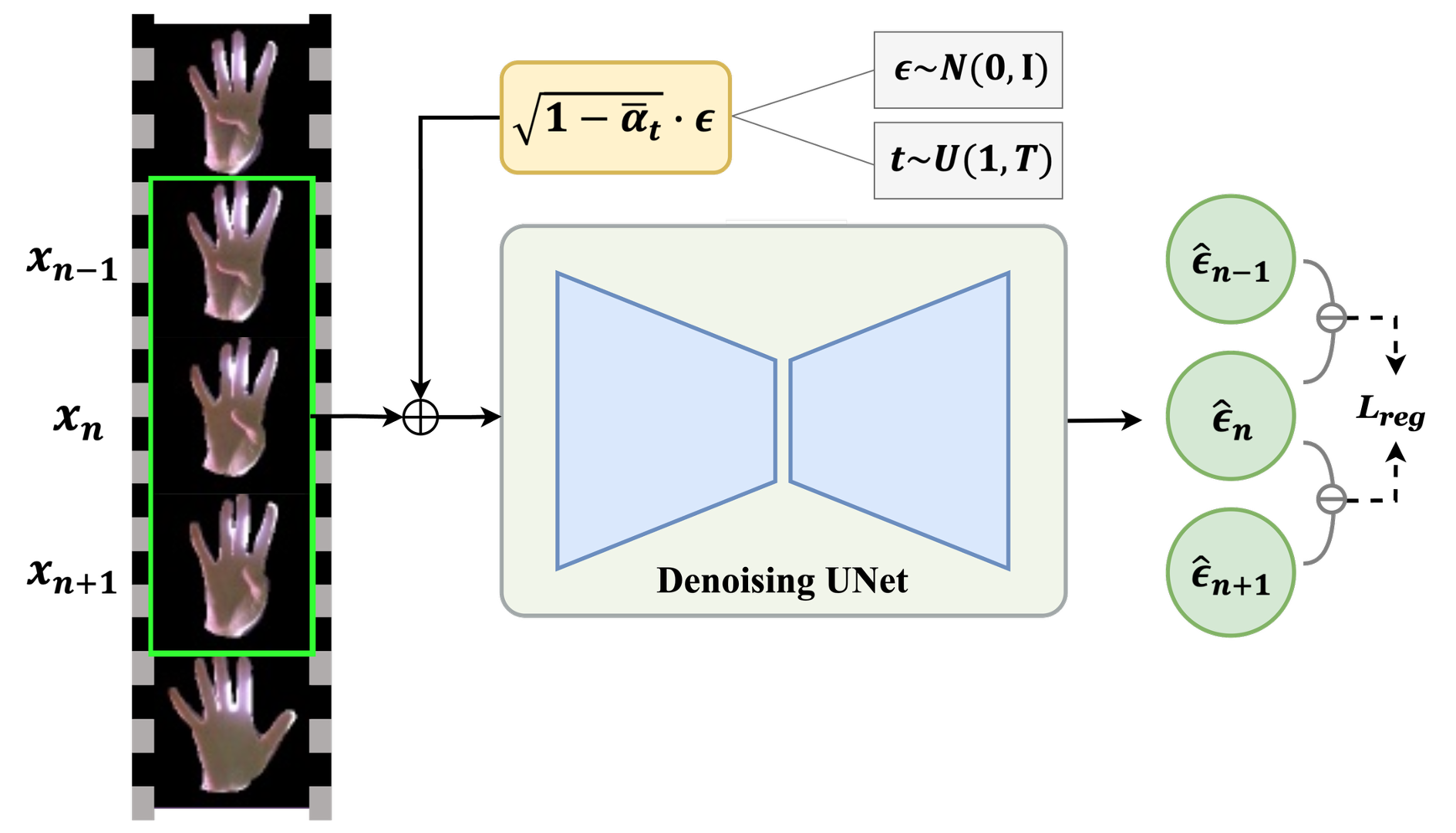}
    \caption{
        Overview of our regularization term.
        Three adjacent video frames are corrupted with shared noise and timestep. A temporal consistency regularization loss is applied to enforce consistency across predicted noise outputs.
    }
    \label{fig:framework}
\end{figure}

Recent studies have begun to examine the generalization ability of diffusion models~\cite{kadkhodaie2023generalization}. Some approaches introduce regularization schemes~\cite{bao2023smooth, xu2023versatile} to improve robustness to unseen samples or encourage smoother latent spaces. Others analyze mode coverage, memorization, and diversity~\cite{aithal2024understanding,karras2022elucidating}, highlighting that diffusion models may still suffer from mode collapse or overfitting under limited or biased training data. Recent representation-learning regularizers provide orthogonal benefits: REPresentation Alignment (REPA) aligns noisy-frame features with pretrained encoder embeddings to speed up diffusion-transformer training~\cite{wang2025diffuse}, while Dispersive Loss repels hidden activations in a positive-free contrastive style, boosting ImageNet FID without extra data~\cite{wang2025diffuse}. Both act purely in latent space, complementing our temporal-proximity regularizer. However, explicit use of temporal information to improve generalization has received limited attention, especially in the context of image diffusion model training. Also, due to the "memorizing" nature of the diffusion objective, regularizing the model to restrain itself from memorizing has direct implications for privacy preservation in diffusion models~\cite{gu2023memorization,hong2024all}.

\section{Method}

Given a collection of static images $\mathcal D_{\text{img}}=\{x_i\}_{i=1}^{N}\subset\mathbb R^{d}$, the empirical measure is the Dirac mixture:
% need to add
\begin{equation}
\mu_{\text{img}}=:
\frac{1}{N}\sum_{i=1}^{N}\delta_{x_i},
\label{eq:dirac_comb}
\end{equation}
where $\delta_{x_i}$ is the Dirac delta distribution centered at sample $x_i$. Consequently, it provides the minimizer with no explicit notion of neighbourhood or smooth variation across data. Under an over‑parameterised network, minimizing i.i.d samples with the diffusion objective encourages memorisation: the model interpolates training samples without regard for their relative arrangement in the ambient space $\mathbb{R}^{d}$, often harming generalisation to unseen data.

Consider a video dataset expressed as continuous trajectories $\{\gamma_v:[0,T_v]\!\to\!\mathbb{R}^{d}\}_{v=1}^{V}$. Integrating Dirac deltas along each path yields the trajectory–supported empirical measure:
\begin{equation}
\mu_{\text{vid}}
=:
\frac{1}{\sum_{v=1}^{V}T_v}
\sum_{v=1}^{V}\int_{0}^{T_v}\delta_{x-\gamma_v(t)},dt,
\label{eq:dirac_field}
\end{equation}
which carries the natural concept of proximity—adjacent frames lie infinitesimally apart along the same curve. Importantly, note that even if one were to train with $\mu_{\text{vid}}$, the exploitation of such a temporal structure is not explicitly encouraged. The interplay between data shuffling in mini-batches and overparameterized networks can still admit memorising solutions. Explicit regularisation is therefore necessary to translate the inductive bias in~\eqref{eq:dirac_field} into well-generalizing smoother hypotheses.

\subsection{Problem Definition}

Let $(x_1, x_2, \ldots, x_N) \in \mathbb{R}^{N \times C \times H \times W}$ denote a sequence of $N$ consecutive video frames, where each frame $x_i$ is a clean image with $C$ channels and a spatial resolution of $H \times W$. For each target frame $x_n$, we define a temporal window $(x_{n-1}, x_n, x_{n+1})$ consisting of the frame itself and its adjacent frames in time. The goal of this study is to train a model that leverages this local temporal context to generate temporally consistent predictions for the target frame $x_n$.

We adopt the standard noise prediction formulation used in denoising diffusion probabilistic models (DDPMs)~\cite{ho2020denoising}. During training, we perform the forward diffusion process on clean images $x \sim p_{\text{data}}(x)$ using a randomly sampled timestep $t \sim \text{Unif}(1, T)$ and Gaussian noise $\epsilon \sim \mathcal{N}(0, I)$:

\begin{equation}
\tilde{x}_t = \sqrt{\bar{\alpha}_t} x + \sqrt{1 - \bar{\alpha}_t} \epsilon,
\end{equation}
where $\bar{\alpha}_t$ denotes the cumulative product of noise scaling coefficients. The denoising model $\epsilon_\theta$ is trained to predict  the added noise using a mean squared error loss~\cite{ho2020denoising}:

\begin{equation}
\mathcal{L}_{\text{mse}} = \mathbb{E}_{x, \epsilon, t} \left[ \left\| \epsilon - \epsilon_\theta(\tilde{x}_t, t) \right\|_2^2 \right].
\end{equation}

However, this basic objective treats each frame independently and fails to directly capture local frame density. To overcome this limitation, we propose a training strategy that exploits temporal information within the triplet frame window to enforce consistency at the noise level.

\subsection{Shared Epsilon-Noise Injection}

% To ensure comparable stochastic conditions across temporally adjacent frames, we introduce a shared noise scaling strategy~\cite{Bao24,Song23}. Specifically, we sample a single diffusion timestep $t \sim \text{Unif}(1, T)$ and compute its corresponding scaling factor $\bar{\alpha}_t$, which is shared across all three frames in the input triplet. While the same noise strength $\bar{\alpha}_t$ is applied, each frame is corrupted using the same Gaussian noise $\epsilon \sim \mathcal{N}(0, I)$. This design enforces a unified perturbation magnitude over time, allowing the model to learn noise predictions that are directly comparable between frames. As a result, it becomes possible to explicitly regularize the model toward temporal coherence in the latent noise space, without requiring architectural modifications such as temporal attention.

% A shared timestep $t$ and noise $\epsilon$ are used to construct noisy versions of the input triplet $(x_{n-1}, x_n, x_{n+1})$:
% \begin{equation}
% \tilde{x}_t^k = \sqrt{\bar{\alpha}_t} \cdot x_k + \sqrt{1 - \bar{\alpha}_t} \cdot \epsilon, \quad \text{for } k \in \{n-1, n, n+1\},
% \end{equation}
% These noisy frames are then independently processed by a denoising model (e.g., U-Net~\cite{Ronneberger15}):
% \begin{equation}
% \hat{\epsilon}_\theta^k = \epsilon_\theta(\tilde{x}_t^k, t), \quad \text{for } k \in \{n-1, n, n+1\}.
% \end{equation}
% This shared-scaling scheme facilitates temporal alignment in the noise prediction space~\cite{Ma24}.

The vanilla diffusion objective perturbs every training frame independently, drawing both the diffusion timestep $t$ and the Gaussian noise $\epsilon$ afresh for each image. That independence makes adjacent frames unrelatable: even if two inputs depict almost identical content, the network observes them through distinct noise realisations at different noise levels.
To this end, before any consistency term can be imposed, we align the stochastic conditions across a local window of K consecutive frames. Concretely we sample
\begin{equation}
\tau \sim \text{U}(0,T), 
\quad 
\boldsymbol{\varepsilon}\sim\mathcal N(\mathbf 0,\mathbf I),
\end{equation}
once per window and inject the same pair $(\tau,\boldsymbol{\varepsilon})$ into every frame $x_i$ in that window:
\begin{equation}
\tilde x^{i} \;=\;\sqrt{\bar\alpha_{\tau}}\,x^{i}\;+\;\sqrt{1-\bar\alpha_{\tau}}\;\boldsymbol{\varepsilon},
\qquad i=1,\dots,K.
\end{equation}
This strategic adjustment leaves the architecture untouched and introduces no extra parameters, yet it synchronises the noise signature and its magnitude across neighbouring frames. Under this aligned structure, we gain access to apply intra-sample objectives. 

% To ensure comparable stochastic conditions across temporally adjacent frames, we introduce a shared noise scaling strategy~\cite{Bao24,Song23}. Specifically, we sample a single diffusion timestep $t \sim \mathrm{Unif}(1, T)$ and compute its corresponding scaling factor $\bar{\alpha}_t$, which is shared across all $N$ frames in the input sequence. While the same noise strength $\bar{\alpha}_t$ is applied, each frame is corrupted using the same Gaussian noise $\epsilon \sim \mathcal{N}(0, I)$. This design enforces a unified perturbation magnitude over time, allowing the model to learn noise predictions that are directly comparable between frames. As a result, it becomes possible to explicitly regularize the model toward temporal coherence in the latent noise space, without requiring architectural modifications such as temporal attention.

% A shared timestep $t$ and noise $\epsilon$ are used to construct noisy versions of the input sequence $(x_1, x_2, \ldots, x_N)$:
% \begin{equation}
% \tilde{x}_t^k = \sqrt{\bar{\alpha}_t} \cdot x_k + \sqrt{1 - \bar{\alpha}_t} \cdot \epsilon, \quad \text{for } k \in \{1, \ldots, N\},
% \end{equation}

% These noisy frames are then independently processed by a denoising model (e.g., U-Net~\cite{Ronneberger15}):
% \begin{equation}
% \hat{\epsilon}_\theta^k = \epsilon_\theta(\tilde{x}_t^k, t), \quad \text{for } k \in \{1, \ldots, N\}.
% \end{equation}

% This shared-scaling scheme facilitates temporal alignment in the noise prediction space~\cite{Ma24}.

\subsection{Temporal Consistency Regularization}

Now, we wish to explicitly inject the geometric structure of the video measure $\mu_{\mathrm{vid}}$ into the learning objective, which should carry an intrinsic notion of proximity along each trajectory. 
We introduce a generic, symmetric, positive weighting function 
$w_{i,j}   \;{=}\;  \varphi\!\;\bigl(\pi_{i,j}\bigr)$
that maps any scalar proximity measure
$\pi_{i,j}\!\in\!\mathbb{R}_{\ge0}$ between frame
$x^{i}$ and $x^{j}$ into a Laplacian weight.
Throughout the paper
$\varphi$ is taken to be a monotone decay function
($\varphi'\!<\!0$), so that two frames judged closer
receive a larger coupling.

Section 3.2 introduced the shared $(\tau,\varepsilon)$ injection so that each frame in a window is observed under identical corruption. This design equalises scale and stochasticity across the window, letting any inter-frame operation possible—an essential pre-condition for our regularization term. Namely, given the aligned noisy inputs $\tilde{x}^{i}$ and their noise predictions $\hat{\varepsilon}_{\theta}^{i}=\varepsilon_{\theta}(\tilde{x}^{i},\tau)$, we penalize disparities between adjacent predictions:

\begin{equation}
\mathcal{L}_{\text{reg}} =  \sum_{i=0}^{N-2} w_{i, j} \left\| \hat{\epsilon}_\theta^{i} - \hat{\epsilon}_\theta^{j} \right\|_2^2,
\end{equation}
with $j=i+1$. The weight $w_{i,i+1}$ acts as a discrete, flow-aware Laplacian, turning $\mathcal{L}_{\mathrm{reg}}$  into a proximity-weighted Dirichlet energy that directly encodes the neighbourhood structure implicit in $\mu_{\mathrm{vid}}$. Finally, our training loss is the composite objective
\begin{equation}
% \boxed{
\mathcal{L}_{\mathrm{total}}
=
\mathcal{L}_{\mathrm{mse}}
\;+\;
\lambda\,\mathcal{L}_{\mathrm{reg}},
% }
% \tag{8}
\end{equation}
Next, we concretize two choices for $\pi_{i,i+1}$
% where $\omega_{i, i+1}^{(b)}$ is a weighting factor computed based on the average optical flow magnitude between frames $i$ and $i+1$, following~\cite{Teed20}:

% with $F^{(b)}_{i \rightarrow i+1}(p)$ representing the optical flow vector at pixel $p$ from frame $i$ to $i+1$ in the $b$-th sample, and $\delta$ being a small constant for numerical stability.

% This flow-guided weighting is highly important in our training, as it dictates the notion of proximity, which is inexistent in standard i.i.d dataset structures.

% To gradually introduce temporal regularization without interfering with early-stage denoising, we apply a linear annealing schedule on $\lambda_{\text{reg}}$ over training step $s$:
% \begin{equation}
% \lambda_{\text{reg}}(s) = 
% \begin{cases}
% \lambda_{\text{start}}, & s < s_{\text{start}} \\
% \lambda_{\text{start}} + \frac{s - s_{\text{start}}}{s_{\text{end}} - s_{\text{start}}} (\lambda_{\text{end}} - \lambda_{\text{start}}), & s_{\text{start}} \leq s < s_{\text{end}} \\
% \lambda_{\text{end}}, & s \geq s_{\text{end}}
% \end{cases}
% \end{equation}

% This schedule allows the model to initially focus on learning the denoising task, and then gradually strengthen the temporal constraint as training progresses. The final objective becomes:

\subsection{Optical-Flow Proximity}\label{sec:flow-weight}
Following classical motion analysis, we set
$\pi_{i,j}$ to the mean-squared optical-flow
magnitude between consecutive frames:
\begin{equation}
  \pi_{i,j}^{flow}
  =
  \frac{1}{HW}
  \sum_{p}\!
  \bigl\|
    F_{i\rightarrow j}(p)
  \bigr\|_2^{2},
\end{equation}
% \begin{equation}
% \omega_{i, i+1}^{(b)} = \left( \frac{1}{HW} \sum_{p} \left\| F_{i \rightarrow i+1}^{(b)}(p) \right\|_2^2 + \delta \right)^{-1}
% \end{equation}
where $F$ is any off-the-shelf flow estimator and $p$ indexes
pixels.  Choosing
$
  \varphi(\pi)=\bigl(\pi+\delta\bigr)^{-1}
$
with a small $\delta\!>\!0$ produces the weight of
Eq.~(7), which serves as a data-driven proxy for geodesic distance along the trajectory manifold~$\mu_{\text{vid}}$.

\subsection{Trajectory Divergence Proximity}\label{sec:dist-deriv-weight}
We consider an estimator-free alternative that
relies solely on the model’s own forward process. The presentation of such is to both highlight a computationally friendly approach to our regularization and convince the flexibility of the design of the proximity.

Let
$
  d^{i,j}
  =
  \frac{1}{CHW}
  \bigl\|
    x_{t}^{i}-x_{t}^{j}
  \bigr\|_2^{2}
$
be the normalised squared distance between two noisy
frames obtained at the same diffusion step~$t$.
We approximate its time derivative by a symmetric
finite-difference:
\begin{equation}
  \pi_{i,j}^{div}=:\dot d^{i,j}
  \;\approx\;
  \frac{
    d_{t+\Delta t}^{i,j} - d_{t-\Delta t}^{i,j}
  }{
    2\,\Delta t
  },
  \label{eq:d-dot}
\end{equation}
with a small integer step
$\Delta t$.  Operationally, the four noisy realisations
$x_{t\pm\Delta t}^{i,j}$ are obtained in a single forward process call, making the computation
lightweight. Intuitively, $\lvert\dot d\rvert$ measures the rate in which the
two trajectories diverge (or converge) in latent space:
frames whose distance changes slowly are deemed proximal.  As for the weighting function, we set
$w_{i,j}
  \;=\;
  \frac{
    1
  }{
    \varepsilon + \lvert \pi_{i,j}\rvert^{\frac{1}{2}}
  }.$

\section{Optimization Analysis}
\label{sec:theory}

For simplicity of our following derivations, we introduce additional notations. Model each training sample as three consecutive frames
$x_0,x_1,x_2$ and denote their UNet noise–predictions by the
$d$-dimensional vectors
$ f_i:=\epsilon_\theta(x_i)\in\mathbb R^{d}$
($d=C\!\times\!H\!\times\!W$).
For any parameter coordinate $\theta$ we write the corresponding
Jacobian rows
$J_i:=\partial_\theta \epsilon_\theta(x_i)\in\mathbb R^{d}$.
Edge‐wise quantities are
$s_{ij}:=\ f_i-f_j$
(aka. output mismatch) and
$D_{ij}:=J_i-J_j$. Now, our temporal regularizer is a discrete Dirichlet energy on the output matching 
\begin{equation}
\label{eq:Es}
E_S(\theta) \;:=\; \frac{1}{2}\sum_{(i,j)\in E} w_{ij}\,\|s_{ij}\|_2^2,
\qquad
\mathcal{L}_{\text{reg}}(\theta)=E_S(\theta),
\end{equation}
Similarly, we can define the Dirichlet energy on the Jacobian matching level 
\begin{equation}
\label{eq:Eg}
E_G(\theta) \;:=\; \frac{1}{2}\sum_{(i,j)\in E} w_{ij}\,\|D_{ij}\|_F^2,
\end{equation}
which measures the roughness of the parameter-space sensitivity field over the local input graph. Let $\ell_i$ be the per-sample denoising loss at $x_i$. A first-order expansion gives the exact decomposition
\begin{equation}
\label{eq:grad-decomp}
\nabla_\theta \ell_i - \nabla_\theta \ell_j \;=\; J_i^\top s_{ij} \;+\; D_{ij}^\top f_j.
\end{equation}
\paragraph{Assumption (Uniform bounds on the local input set).}
Let $\mathcal{N}:=\{x_i\}_{i=1}^N$ denote the neighborhood/window used to form $E$.
There exist finite constants $G,F>0$ such that
\[
\sup_{x\in \mathcal{N}} \big\|\tfrac{\partial f(x,\theta)}{\partial\theta}\big\|_{2} \le G
\quad\text{and}\quad
\sup_{x\in \mathcal{N}} \|f(x,\theta)\|_{2} \le F.
\]
With these pre-set bounds, we can contain the difference norm of adjacent gradients in Eq.  (\ref{eq:grad-decomp}) as
\begin{equation}
\label{eq:pairwise}
\|\nabla_\theta \ell_i - \nabla_\theta \ell_j\|_2^2
\;\le\; 2G^2\,\|s_{ij}\|_2^2 \;+\; 2F^2\,\|D_{ij}\|_2^2.
\end{equation}
\begin{figure}[b]
    \centering
    \includegraphics[width=.85\linewidth]{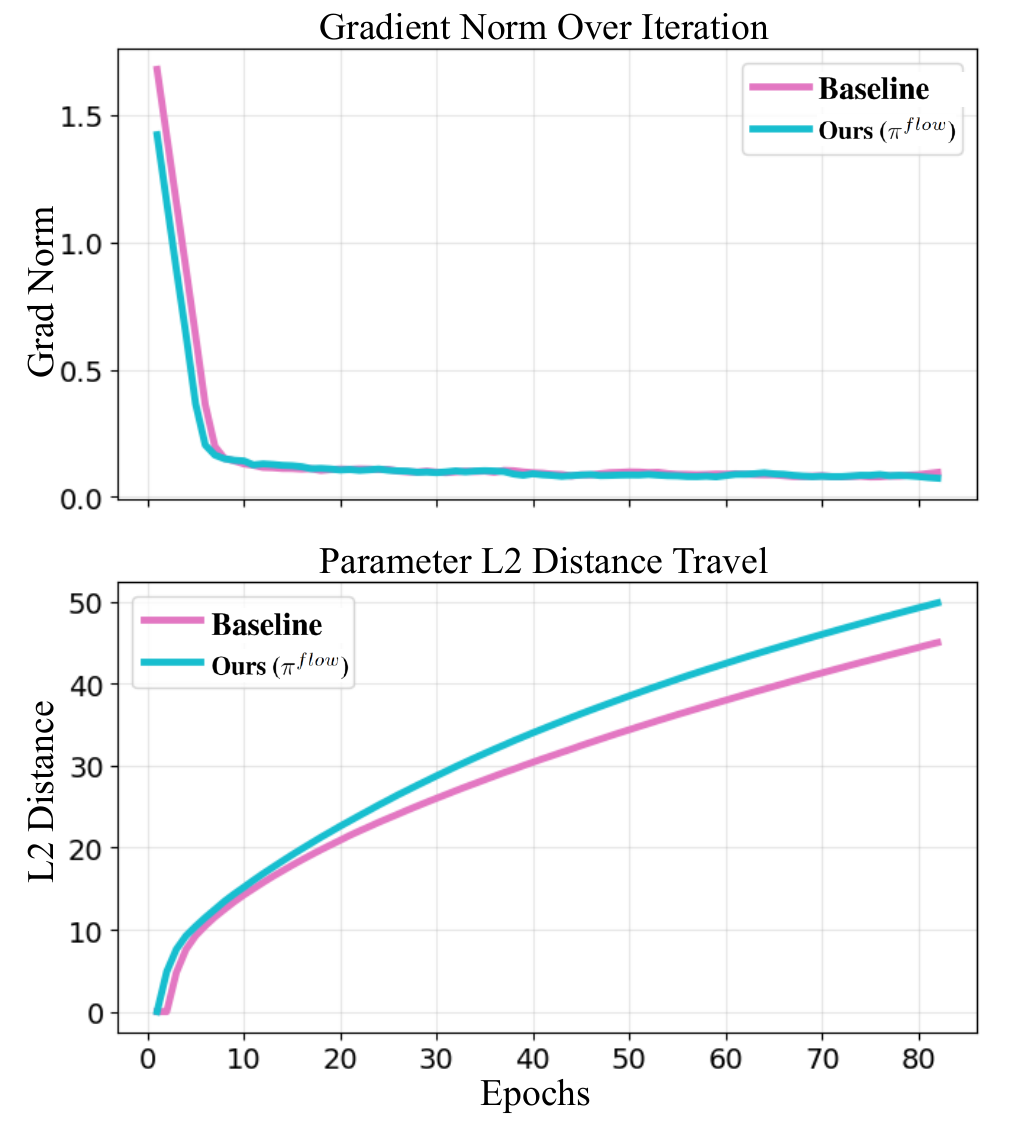}
    \caption{Gradient norm and $L^2$ parameter distance, compared across baseline and our ($\pi^{flow}$ variant) model.
    }
    \label{fig:travel}
\end{figure}
In other words, jointly matching outputs and Jacobians suppresses per-sample gradient disagreements. While the disagreement $\|s_{ij}\|$ is explicitly minimized by enforcing regularization, the first-order approximation of the updated Jacobians is  
\begin{equation}
J_i^{+}\approx J_i-\eta\,H_j\,\nabla_\theta \mathcal L_{\mathrm{reg}}
\end{equation}
Yielding the difference of two adjacent Jacobians $J_i$ and $J_j$, we have
\begin{equation}
D_{ij}^{+}\;\approx\; D_{ij}-\eta \underbrace{(H_i-H_j)}_{\text{frames' curvature gap}}\nabla_\theta \mathcal L_{\mathrm{reg}}
\end{equation}
The update subtracts a component of $D_{ij}$ driven by its correlated portion with the current discrepancy $s_{lm}$. As those discrepancy directions vary across the updates, the regularization continuously remove this very correlation and $||D_{ij}||$ shrinks. 

Now let $L_w$ be the (unnormalized) graph Laplacian on the local graph $(\{1,\dots,N\},E)$ with weights $w_{ij}$, and $\lambda_2(L_w)>0$ its algebraic connectivity (aka. the second smallest eigenvalue). For node-attached vectors $u_i\in\mathbb{R}^P$ with mean $\bar u=\frac{1}{N}\sum_i u_i$, the discrete Poincar\'e inequality states
\begin{equation}
\label{eq:poincare}
\frac{1}{N}\sum_{i=1}^N \|u_i - \bar u\|_2^2
\;\le\;
\frac{1}{N\,\lambda_2(L_w)}\sum_{(i,j)\in E} w_{ij}\,\|u_i - u_j\|_2^2.
\end{equation}
Choosing $u_i=\nabla_\theta\ell_i$ and applying \eqref{eq:pairwise} yields the explicit variance bound
\begin{equation}
\label{eq:var-bound}
\underbrace{\frac{1}{N}\sum_{i=1}^N \|u_i - \overline{u}\|_2^2}_{\mathrm{Var}(\nabla_\theta\ell)}
\;\le\;
\frac{4}{N\,\lambda_2(L_w)}\Big(G^2\,E_S(\theta) \;+\; F^2\,E_G(\theta)\Big),
\end{equation}
where $\overline{u} := \frac{1}{N}\sum_{i=1}^N \nabla_\theta\ell_i$ is the batch-mean (global) gradient. Equation~\eqref{eq:var-bound} formalizes a compact message: local regularization that matches outputs ($E_S$) and enforces Jacobian similarity ($E_G$) over adjacent inputs directly controls the global dispersion of per-sample gradients around their batch mean. Since SGD uses the batch mean as its update direction, reducing $\mathrm{Var}(\nabla_\theta\ell)$ improves the signal-to-noise ratio of the update, producing straighter optimization trajectories and faster, more stable convergence. Empirically, we calculate the average gradient norm for 50 epochs for both the baseline and our model. In tandem, we measure the l2 parameter distance between the updating parameter and its initialization, shown in Fig.~\ref{fig:travel}. Notice that with a comparable gradient norm budget, the model parameter travels farther.

% \begin{figure}
%     \centering
%     \includegraphics[width=.85\linewidth]{figure/travel.png}
%     \caption{Gradient norm and $L^2$ parameter distance, compared across baseline and our ($\pi^{flow}$ variant) model.
%     }
%     \label{fig:convergence}
% \end{figure}

\section{Experiments}

\paragraph{Experimental Setup.} We evaluate our method using the HandCo dataset~\cite{Freihand2019,ZimmermannAB21}, which consists of hand gesture video sequences captured in controlled environments, as it contains abundant temporal variation due to frequent finger and hand articulation over time, making it well-suited for evaluating the impact of temporal supervision. Moreover, the nature of hand motion in video often leads to transient occlusions or anatomically ambiguous frames—such as missing or overlapping fingers—which can challenge static supervision methods and highlight the benefits of temporally informed training. To solely focus on the gesture itself, we mask the background in all frames. All models are trained with the epsilon-prediction objective (DDPM). During sampling, we iterative 100 sampling steps with DDIM~\cite{song2020denoising}. We use a U-Net convolution-attention hybrid backbone for the denoising network across all experiments.

\paragraph{Training Variants.}To assess the benefit of explicit regularisation, we compare across varied configurations. All settings use the same video dataset.
\begin{itemize}[leftmargin=*]
\item \textbf{Baseline}: frames are sampled independently across the entire dataset, obliterating temporal structure.
\item \textbf{Sequence‑preserving}: sampling is i.i.d. at the video level, while frame order within each clip is retained, exposing short trajectories but without coupling across them. This is to encourage noisy variation canceling across similar frames in the batch.
\item \textbf{Adjacent consistency}: We ablate the weighting in our regularization.
\item \textbf{Dispersive Loss}: We apply a representation learning-centric regularization loss, which can be understood as a contrastive loss without penalizing positive terms. For more details, please refer to their work~\cite{wang2025diffuse}.
\item \textbf{Ours ($\pi^{flow}$)}: Our regularization with optical flow-based proximity.
\item \textbf{Ours ($\pi^{div}$)}: Our regularization with trajectory divergence proximity with $\Delta t =50$. 
\end{itemize}

For all baselines, we train a standard DDPM using static single-frame supervision ($seq\_len=1$) with a batch size of 256 for 500 epochs. The learning rate is set to 0.0001, and exponential moving average (EMA) with a decay of 0.9995 is applied. For the dispersive loss, we set the temperature at 0.5, with $\lambda=0.005$
For our methods, we adopt the same architecture and diffusion configuration but extend the input sequence length to three consecutive frames ($seq\_len=3$). We set $\lambda= 0.1$. The batch size is set to 128 and training is conducted for 300 epochs.
Both models are trained with 25,000 samples randomly selected from the HandCo training set. During evaluation, we generate 50,000 images and report results based on FID and qualitative sample diversity. All experiments are conducted using 4 NVIDIA A5000 GPUs.

\subsection{Experimental Results}

% \begin{table}[t]
% \centering
% \begin{tabular}{|l|l|l|}
% \hline
% FID-25k               & Train  & Val     \\ \hline
% baseline       & 4.02 & 11.74 \\ \hline
% ours (w/o opt. flow) &  4.59      &  11.67       \\ \hline
% ours           & \textbf{3.83} & \textbf{10.88} \\ \hline

% \end{tabular}
% \caption{FID scores for baseline and our method. Lower is better. The baseline result corresponds to the best model at 450 training epochs, while our method achieves its best performance at only 150 epochs.}
% \label{tab:fid}
% \end{table}

% \begin{figure}[t]
%     \centering
%     \includegraphics[width=1.0\linewidth]{figure/fidconvergence2.pdf}
%     \caption{FID scores over training iterations for the baseline and our ($\pi^{flow}$ variant) method. Our approach converges faster and consistently achieves lower validation FID, demonstrating improved training efficiency and generalization performance.
%     }
%     \label{fig:convergence}
% \end{figure}

\begin{figure}[t]
    \centering
    \includegraphics[width=1.0\linewidth]{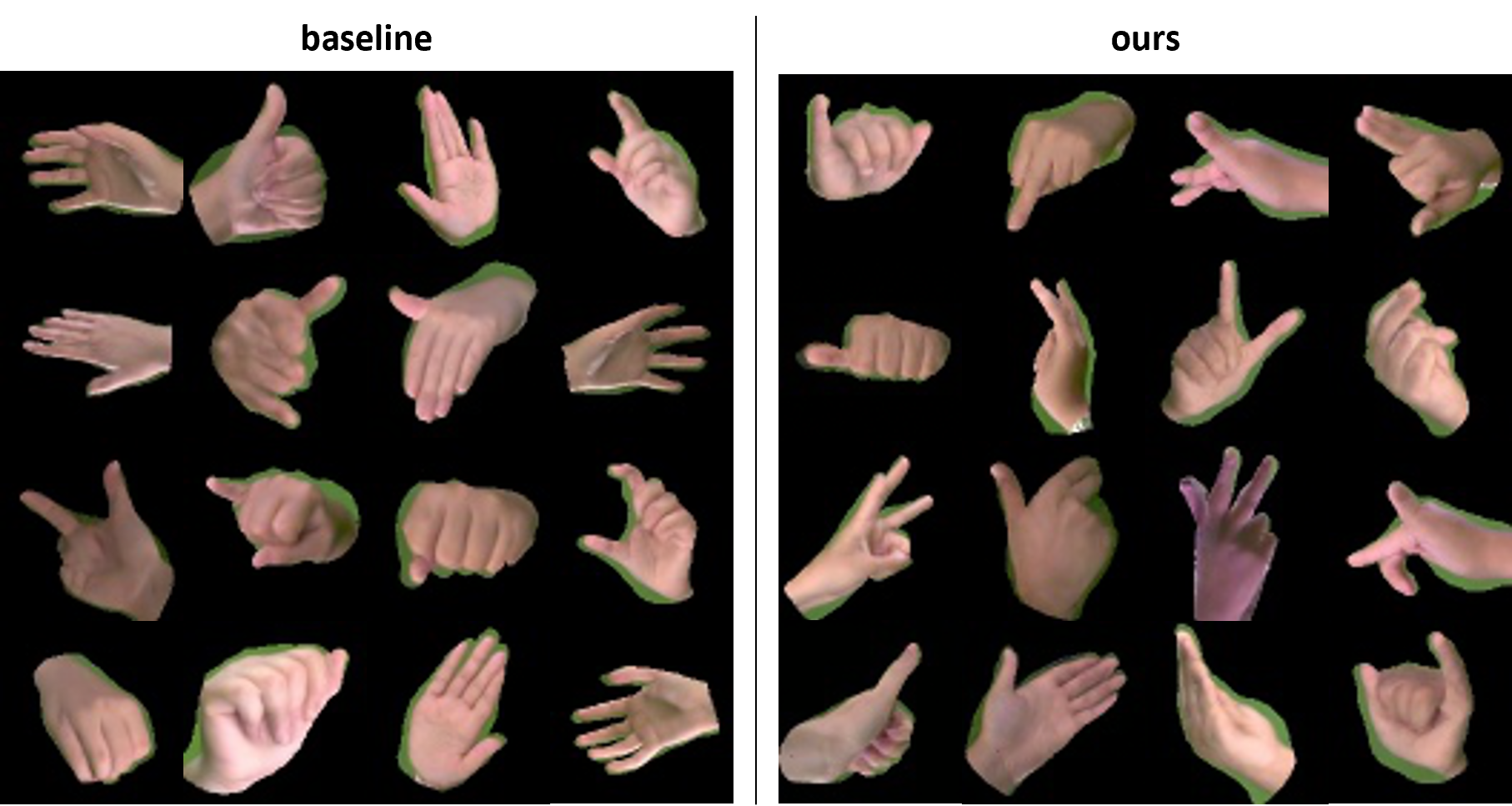}
    \caption{Qualitative comparison of generated images. Left: samples from the baseline model. Right: samples generated by our ($\pi^{flow}$ variant) method. Our model produces more diverse outputs.
    }
    \label{fig:sample}
\end{figure}
% \begin{figure}[t]
%     \centering
%     \includegraphics[width=1.0\linewidth]{figure/figure_last.png}
%     \caption{FID scores over training iterations for the baseline and our method. Our approach converges faster and consistentlFrom isolated impulses to a Temporally-Integrated Dirac Fieldy achieves lower validation FID, demonstrating improved training efficiency and generalization performance.
%     }
%     \label{fig:convergence}
% \end{figure}

% table 최종 best 만 보여줄 것 
% witouout flow 추가 (ablation 용도) 
% baseline / ours best epoch 에서 
%sampling step NFE 라고 작성 .... 

\cref{tab:fid} presents the FID scores on both the training and validation sets for the baseline and our proposed method. The baseline reaches its best performance at 450 training epochs, whereas our method achieves superior FID scores after only 150 epochs. This highlights that our temporally grounded training strategy not only enhances generative quality and generalization but also accelerates convergence.
Furthermore, as illustrated in \cref{fig:convergence}, our method achieves comparable or better validation FID scores approximately 2.5× faster than the baseline, clearly demonstrating its training efficiency and effectiveness.

\begin{table}[t]
\centering
\renewcommand{\arraystretch}{0.8}
\setlength\tabcolsep{16.0pt}
\resizebox{1.\linewidth}{!}{
\begin{tabular}{c|cc}
\toprule
FID-25k & Train  & Val  \\
\midrule
Baseline & 4.02 & 11.74 \\
Seq. Preserving & 3.98 & 11.70 \\
Adj. Consistency & 4.59 & 11.67 \\
Dispersive Loss & 4.68 & 12.25 \\
Ours ($\pi^{flow}$) & \underline{3.31} & \underline{11.23} \\
Ours ($\pi^{div}$) & \textbf{3.02} & \textbf{10.87} \\
\bottomrule
\end{tabular}}
\caption{FID scores for baseline and our method. Lower is better. We report performance of models (at their best) at 450-500 training epochs, while our method achieves its best performance at early as 150 epochs.}
\label{tab:fid}
\end{table}
\cref{fig:sample} presents a qualitative comparison between the baseline model and our proposed method. While the baseline model tends to produce visually similar or repetitive outputs, our model generates a broader range of appearances with greater variation in pose, shape, and orientation. This suggests that the proposed temporal supervision strategy encourages the model to fairly treat samples of varying modes equally. 

\section{Discussion}
%wo flow guided weighting 결과 설명 
\subsection{Ablation Studies}
We study the convergence behavior by ablating and highlighting exclusive components. Namely, we notice that the preservation of the sequence during parameter update incrementally helps on convergence. As hypothesized, we believe this advantage arises from the averaging of noisy gradients per sequence, resulting in higher fidelity sequence-wise gradients. However, even with such a structured sampling, we see that it is not sufficient for any significant improvement over the baseline. For the case of adjacent consistency, where we ablate the proximity weighting, the model suffers noticeable performance degradation: compared to ours ($\pi^{div}$), the validation FID increases from 10.87 to 11.67, and the training FID from 3.02 to 4.59. This confirms that flow-guided weighting plays a crucial role in offering the right Laplacian weighting matrix. Interestingly, Dispersive Loss doesn't have a clear positive contribution, both in train and validation FID. Lastly, we verify that  while both designs achieve superior performances,  $\pi^{div}$) show stronger model fitting (Train FID) and generalization (Validation FID).

 \subsection{Limitations}
 
Our formulation presumes that consecutive frames trace smooth trajectories on the data manifold (Sec. 3.3). In settings with abrupt shot changes, strong camera shake, or heavy motion blur, the optical-flow magnitudes or the empirical trajectory divergence used to weight the Laplacian can become unreliable, leading either to vanishing weights or—in the worst case—to spurious penalties that harm convergence. Robust flow estimation—or alternative self-supervised proximity signals—will be required before the method can be safely deployed on unconstrained “in-the-wild” video. In regards to the optimization analysis, the variance-reduction argument assumes idealized training conditions (e.g. bounded Hessians and Polyak-Lojasiewicz (PL) Condition). While the empirical trends support the analysis, stronger guarantees remain open.

% \subsection{Selective Alignment}
% Note that only the component of $d_{ij}$ parallel to $s_{ij}$
% appears in Eq.\,(9).  Orthogonal Jacobian differences
% $d_{ij}^{\perp}$ incur zero increase in $R(\theta)$ and are therefore blind to the objective. Alternatively, one could have imposed a direct Jacobian alignment loss. However, such an alignment may be overly strict on the Jacobians' expressivity. Also, calculating Jacobians during training slows down training severely, defeating the purpose of faster convergence.
% \section{Conclusion}

% We present a simple yet effective training strategy for image diffusion model that leverages temporal information from real video sequences. Our method injects shared noise across consecutive frames and applies a temporal consistency regularization in the noise prediction space, encouraging the model to align its prediction over nearby frames. We provide theoretical analysis showing that our regularization ultimately reduces gradient variance and accelerates convergence by promoting Jacobian alignment across temporally adjacent frames. Experiments on the Handco dataset validate that our method leads to faster and better convergence, lower FID scores on both training and validation datasets, and more diverse outputs, all without requiring architectural modifications. Theses findings highlight the potential of temporal supervision as a general principle for improving generalization in image diffusion models. 

\section{Conclusion}
We introduced a proximity-based regulariser for diffusion models that ties together the denoising predictions of neighbouring frames under shared noise by using temporally structured video datasets.  The weight $\omega$ is deliberately agnostic to any one metric: it can be driven by optical-flow magnitude or, as we show, by a new latent trajectory divergence that looks at how inter-frame distance changes along the diffusion path.  When applied to hand-gesture data, this “frame-bridging” constraint forces the generator to respect the subtle pose shifts and finger articulations that separate one gesture phase from the next, improving both visual fidelity and temporal coherence.  The approach requires no external trackers, adds negligible cost, and offers a principled template for injecting fine-grained temporal semantics into generative training. Empirically, its effectiveness is pronounced in both train FID and validation FID through better fitting without overfitting.

\section*{Acknowledgments}
We acknowledge the help and contribution of the following people: Yulim So for the optical flow calculation, method section writing, and diagram drawing, Sehun Chang for plotting the results, Soobin Cha for reviewing the paper, and Jae-Pil Heo for providing the guidance for the project. This work was partly supported by Institute for Information \& communication Technology Planning \& evaluation (IITP) grants funded by the Korean government MSIT:
(RS-2022-II221199, RS-2022-II220688, RS-2019-II190421, RS-2023-00230337, RS-2024-00356293, RS-2024-00437849, RS-2021-II212068,  RS-2025-02304983, and RS-2025-02263841).

{
    \small
    \bibliographystyle{ieeenat_fullname}
    \bibliography{main}
}

\end{document}